\documentclass[conference]{IEEEtran}
\IEEEoverridecommandlockouts
\usepackage{cite}
\usepackage{amsmath,amssymb,amsfonts}
\usepackage{algorithmic}
\usepackage{graphicx}
\usepackage{textcomp}
\usepackage{xcolor}
\usepackage{amsmath}
\usepackage{amssymb}
\usepackage{booktabs}
\usepackage{hyperref}
\usepackage{lipsum}
\usepackage{caption}
\usepackage{indentfirst}

\def\BibTeX{{\rm B\kern-.05em{\sc i\kern-.025em b}\kern-.08em
    T\kern-.1667em\lower.7ex\hbox{E}\kern-.125emX}}
\begin{document}

\title{LLM-Enabled Style and Content Regularization for Personalized Text-to-Image Generation \\
}

\newcommand{\myrefmark}[1]{\textsuperscript{#1}}
\author{%
\centering

\IEEEauthorblockN{
                  Anran Yu\textsuperscript{1\,\textasteriskcentered},~~
                  Wei Feng\textsuperscript{1\,\textasteriskcentered},~~
                  Yaochen Zhang\textsuperscript{3},
                  Xiang Li\textsuperscript{1},~~
                  Lei Meng\textsuperscript{1,2}\IEEEauthorrefmark{4} ,~~
                  Lei Wu\textsuperscript{1}\IEEEauthorrefmark{4},~~
                  Xiangxu Meng\textsuperscript{1}~~
                  }

\vspace{0.2em}

\IEEEauthorblockA{%
\textit{\textsuperscript{1} School of Software, Shandong University, Jinan, China} \\
\textit{\textsuperscript{2} Shandong Research Institute of Shandong University, Jinan, China} \\
\textit{\textsuperscript{3} Inspur Technology, Jinan, China} \\
\textit{Emails: (y\_ar, 202200300383, xiangli\_)@mail.sdu.edu.cn, (lmeng, i\_lily, mxx)@sdu.edu.cn, zyc@inspur.com} \\
}
}
\renewcommand{\thefootnote}{\fnsymbol{footnote}}
\twocolumn[{
\renewcommand\twocolumn[1][]{#1}
\maketitle
\begin{center}
    \captionsetup{type=figure}
    \includegraphics[width=1\textwidth]{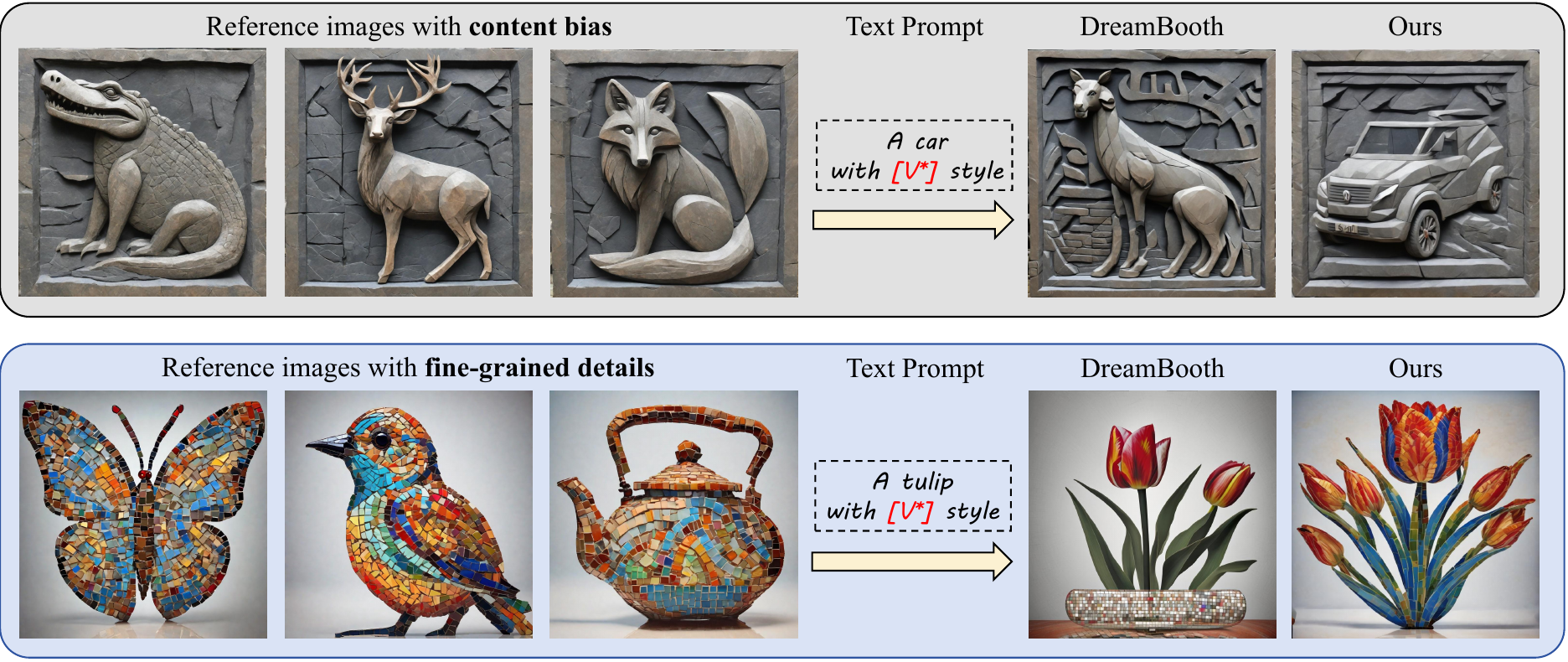}
    \captionof{figure}{\textbf{Comparison with DreamBooth in the task of personalizing text-to-image generation.} The first line features style images with content bias, where the car generated by DreamBooth resembles an animal, as the reference images are all animals, whereas our method accurately generate the car. The second line showcases style images with fine-grained details, the tulip generated by DreamBooth demonstrates insufficient stylization, while our approach produces a richly detailed stylized tulip. The text prompt follows the format of ``an obj with [V*] style'', where [V*] represents the style identifier, which associated with the style of the reference images.}
\vspace{6pt}
\label{FigHead}
\end{center}
}]

\begin{abstract}
The personalized text-to-image generation has rapidly advanced with the emergence of Stable Diffusion. Existing methods, which typically fine-tune models using embedded identifiers, often struggle with insufficient stylization and inaccurate image content due to reduced textual controllability. In this paper, we propose style refinement and content preservation strategies. The style refinement strategy leverages the semantic information of visual reasoning prompts and reference images to optimize style embeddings, allowing a more precise and consistent representation of style information. The content preservation strategy addresses the content bias problem by preserving the model's generalization capabilities, ensuring enhanced textual controllability without compromising stylization. Experimental results verify that our approach achieves superior performance in generating consistent and personalized text-to-image outputs.

\end{abstract}

\begingroup
\vspace{0.2em}
\hrule
\renewcommand{\thefootnote}{}
\footnotetext{* Equal contribution; junior author listed first.}
\footnotetext{§ Corresponding author.}
\endgroup



Text-to-image models, a subset of generative models, have rapidly evolved to convert natural language descriptions into corresponding visual representations, as demonstrated in recent works \cite{xu2024imagereward, kim2023dense, kang2023scaling, shi2024instantbooth, hao2024optimizing}. Concurrently, personalized image generation has achieved notable progress, enabling the production of images that align with user preferences such as aesthetic styles, color schemes, thematic choices, and specific artistic styles. For widely recognized artistic styles, such as Vincent van Gogh’s works or common styles like watercolor and sketch, users can effectively integrate these descriptors into text prompts, and pre-trained models can generate outputs that faithfully capture their distinct characteristics. However, describing rare or ambiguously defined artistic styles using general textual expressions remains a significant challenge, as these styles often encompass intricate nuances in color palettes, textures, and other defining features \cite{huang2024creativesynth, liu2024museummaker, park2024text}. Although this challenge can be avoided by directly inputting stylized reference images, existing personalizing text-to-image generation models still suffer from pronounced overfitting issues, compromising content consistency and leading to subpar visual results. Furthermore, the process of stylization often affects the underlying image content, leading to a mismatch between the generated image and the input textual prompt. This occurs because models tend to prioritize stylistic elements, such as color schemes or textures, at the expense of accurately representing the described content. Consequently, the generated images may diverge from the textual input, with objects, scenes, or details being inaccurately depicted or entirely omitted, thereby exacerbating the challenge of achieving both content accuracy and faithful representation of diverse artistic styles.

To address these limitations, researchers have explored methods like `prompt optimization’ \cite{kim2024selectively, lee2024direct}, which leverage Visual Language Models (VLMs) to optimize textual prompts and better capture stylistic features. However, the effectiveness of these approaches is limited by the optimization capabilities of VLMs, often leading to variable outcomes. To improve consistency and coherence, some studies \cite{ruiz2023dreambooth, shi2022loss, liu2024prior} employ prior loss functions, which align generated outputs with intended content by penalizing deviations from prior knowledge of the target subject. While this ensures consistency across generated images and preserves diversity, prior loss focuses on specific object fidelity rather than enhancing the general diversity of diffusion model outputs, limiting its application in broader style representations.

To tackle these challenges, we propose a method for personalizing text-to-image generation to improve image stylization and text alignment. Specifically, our approach begins by extracting image style keywords using Visual Language Model reasoning, which are then used to initialize the embeddings of the style identifiers (e.g., \(V^*\)). These embeddings are further fine-tuned to achieve a more precise and semantically rich representation of the desired style. Following this initialization, we simultaneously fine-tune the attention layers in both the U-Net and Text Encoder, enabling the model to better capture specific styles and improve its understanding of text descriptions. This joint optimization enhances the alignment between text prompts and generated images while preserving stylistic coherence. Additionally, we introduce a content preservation prior loss function to ensure the generated images maintain accurate content consistency with the input prompts. Together, these components form a comprehensive framework that mitigates overfitting, balances style and content consistency, and achieves superior performance in personalized style generation tasks.

Our contributions are as follows:
\begin{itemize}




    \hyphenpenalty=5000
    \emergencystretch=3em
    \item We propose a personalized text-to-image generation me-thod  based on Visual Language Model reasoning, achieving more accurate initialization of style embeddings and improved representation of stylistic details. 
    

    \item We introduce a content preservation strategy with a content preservation prior loss to address overfitting and content inconsistency challenges in personalizing style transfer tasks, effectively balancing content consistency and style adaptation.
    
    \item We evaluate our method using two public datasets, showcasing its effectiveness through superior performance in both style representation and content alignment, underscoring the robustness of our approach.

\end{itemize}

\begin{figure*}
  \centering
  \includegraphics[width=0.95\linewidth]{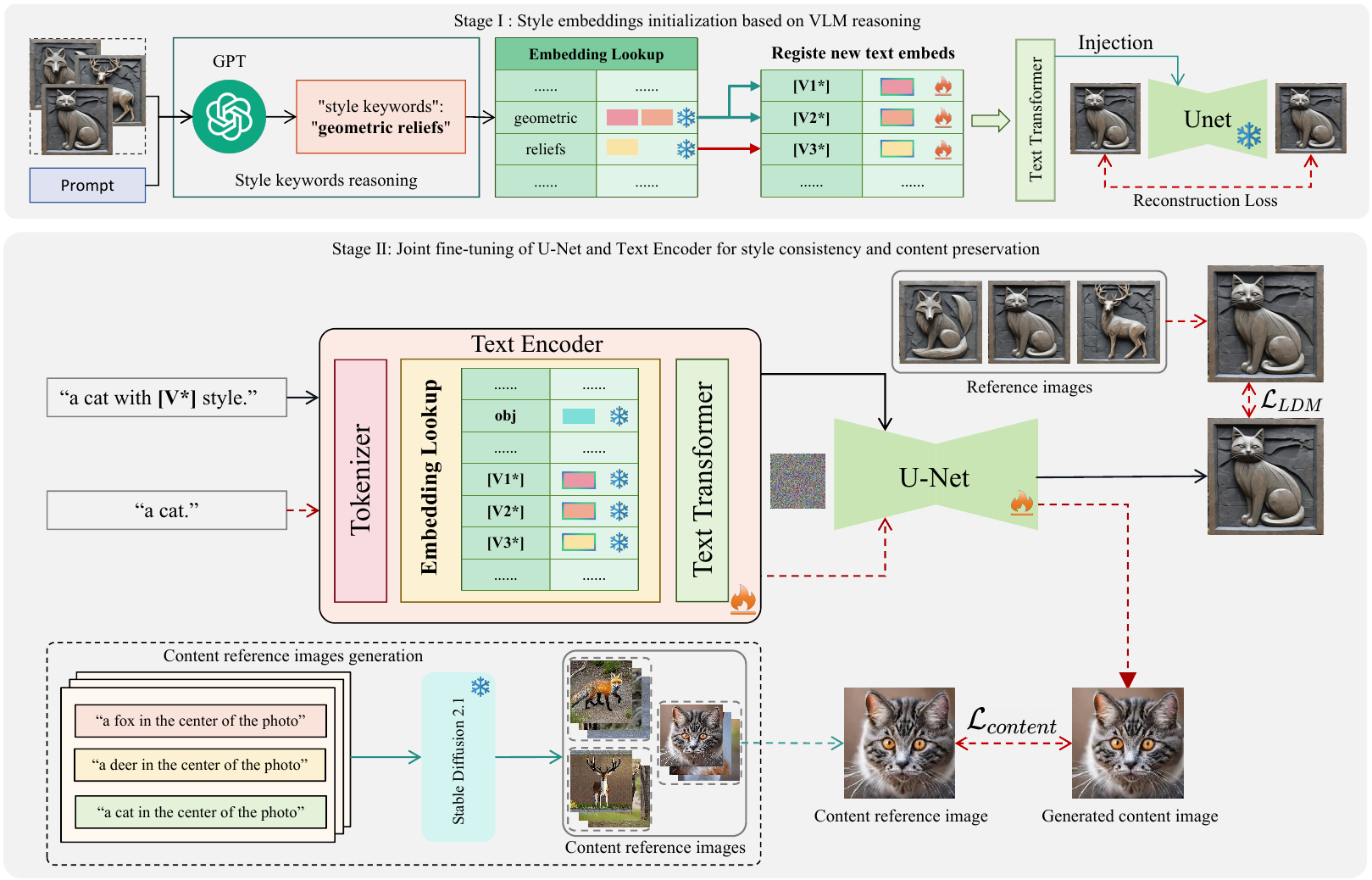}
\caption{\textbf{The overview of our method.} The training consists of two stages. In the first stage (top), a Visual Language Model (e.g., GPT) is guided by a specially designed prompt to generate style-specific keywords of reference images, which are used to initialize new embeddings of style identifiers (e.g., \(V^*\)), while existing embeddings are frozen to preserve semantic integrity. The new embeddings are optimized through reconstruction loss, aligning the outputs with the desired style. In the second stage (bottom), we use the frozen Stable Diffusion 2.1 to generate content reference images for each object in the style category, then object descriptions with and without [V*] style are processed through Text Encoder to obtain latent space codes. These codes are then fed into the U-Net to generate two sets of images. The set with the [V*] style identifier is compared to the reference images to calculating the $L_{\text{LDM}}$, while the other is evaluated against content reference images to calculating the $\mathcal{L}_{\text{content}}$.}
  \label{FigModel}
\end{figure*}


\section{Related Work}
\subsection{Text-to-image diffusion models}
Diffusion models have become a cornerstone in text-to-image synthesis, employing Markov chains to iteratively produce high-quality images from textual descriptions through forward and reverse diffusion steps. In the forward step, noise is added to the image, and in the reverse step, the model learns to denoise and reconstruct the image to match the given conditions. Prominent models such as Stable Diffusion \cite{rombach2022high}, Imagen \cite{saharia2022photorealistic}, and DALL-E 2 \cite{ramesh2022hierarchical} have set benchmarks for generating highly detailed and faithful images. While these models excel at aligning textual inputs with visual outputs, there remains a growing demand for more precise control over stylistic and compositional elements. Methods like ControlStyle \cite{chen2023controlstyle} allow fine-grained manipulation of specific visual attributes, giving users enhanced control, while Region-Aware Diffusion \cite{huang2023region} improves text-driven editing by targeting specific image regions with directional guidance. However, challenges persist in decoupling style and content, leading to style-content confusion and reduced adaptability for unconventional styles or rare contexts. Additionally, these methods may fail to capture intricate details, further limiting their flexibility. Our approach, leveraging Stable Diffusion \cite{rombach2022high}, emphasizes maintaining stylistic consistency and content fidelity, even for rare or complex styles. By addressing irrelevant elements and ensuring alignment between text and visual output, we deliver highly coherent and aesthetically appealing results.

\subsection{Personalized image synthesis}
Personalized image synthesis focuses on extracting distinct concepts from reference images for precise recreation or modification. Recent advances \cite{esser2024scaling, ma2024style,van2023anti, 10447042} have refined high-resolution image generation by introducing innovative methods. Rectified Flow \cite{esser2024scaling} establishes direct connections between data and noise to enhance the generative process, while Style Creation \cite{ma2024style} uses perceptual and texture enhancement loss to merge multiple styles into content images, creating unique, user-tailored results. Attention Injection \cite{10447042} further improves the efficiency and quality of personalized synthesis by optimizing feature handling through enhanced attention mechanisms. Despite these advances, these methods often require extensive computational resources and are heavily reliant on diverse, high-quality training data. Moreover, they frequently struggle with generating sufficient detail, limiting their generalizability and performance. In contrast, our model achieves significant improvements in visual quality and detail while overcoming these limitations. With only 3 style images, our approach delivers highly detailed and stylistically coherent outputs, demonstrating   ability to achieve high performance with limited data availability.


\section{Methods}
\subsection{Problem Definition}

The task of personalizing text-to-image generation is to synthesize an image \( y \) that integrates the semantic intent of a text prompt \( t \) with the stylistic attributes of reference images \( s \). Let \( T = \{ t_k \}_{k=1}^{N_t} \) represent the set of text prompts, and \( S = \{ s_j \}_{j=1}^{N_s} \) denote the set of style reference images. The goal is to learn a mapping function \( f \) such that:
\[
y = f(s, t), \quad \forall s \in S, \, t \in T
\]
where \( f \) synthesizes content from \( t \) and style from \( s \). 

To achieve this, we fine-tune a pre-trained model \( M \) using style reference images \( S \), adapting it to the desired style attributes. After fine-tuning, the model encodes the semantic information of the text prompt \( t \) into embeddings \( V_t \):
\[
V_t = M(t)
\]
The model then combines \( V_t \) with the style features learned during fine-tuning to generate the stylized output \( y \), aligning with both \( s \) and \( t \).

The complexity of this task depends on the coverage of the training set \( Z_{\text{train}} = \{ (s_j, t_k) \}_{j,k} \) within the full style space \( Z_{\text{full}} \). If \( Z_{\text{train}} \approx Z_{\text{full}} \), the task resembles a conventional style transfer problem. However, if \( Z_{\text{train}} \subset Z_{\text{full}} \), the model must generalize to unseen style-content combinations, which increases the difficulty of preserving both content and stylistic fidelity. Additionally, challenges such as embedding entanglement \cite{kim2024selectively}, including biases from backgrounds, nearby objects, or materials, can introduce artifacts, including texture inconsistencies and color mismatches, further complicating the generation process.

\subsection{Model Architecture}
Our model is based on the DreamBooth framework, using the pretrained Stable Diffusion 2.1 model to facilitate personalizing text-to-image generation. As shown in Figure \ref{FigModel}, we introduce style refinement strategy to extract style keywords from reference images based on VLMs, and optimize the embedding lookup in the Text Encoder, improving the representation of style information of the style identifier. During the joint fine-tuning stage, we simultaneously fine-tuned the multi-head attention layers in both the U-Net and Text Encoder of the pre-trained model, further enhancing the model’s adaptability to new styles. To address overfitting, we introduce a content preservation prior loss function. Overall, the training process consists of two stages: a style refinement stage to learn and apply styles from reference images, and a joint fine-tuning stage to preserve the semantic integrity of the input as well as further fine-tune the model, to better generate stylized images with correct content.

\subsection{Style Refinement}
As illustrated in Figure \ref{FigHead}, the original baseline model that uses DreamBooth, shows significant discrepancies in content, color accuracy, and stylistic details. These problems stem primarily from the initialization of text embeddings of style identifier (e.g., \( V^* \)), which often lack robust semantic clarity. Additionally, the diminished controllability of these embeddings arises from the unintended integration of style-irrelevant information—such as descriptors for animals (Figure \ref{FigHead})—leading to a phenomenon known as embedding entanglement. While several approaches have sought to address this entanglement through more nuanced text prompts, the inherent ambiguity of natural language complicates the precise articulation of specific styles, particularly those that are rare or inadequately defined, ultimately resulting in suboptimal stylization outcomes that degrade the visual experience. To address these limitations and improve control over stylistic details, we propose a novel approach for initializing style embeddings that reduces entanglement and enhances precision in style transfer.

\begin{figure*}
  \includegraphics[width=1\linewidth]{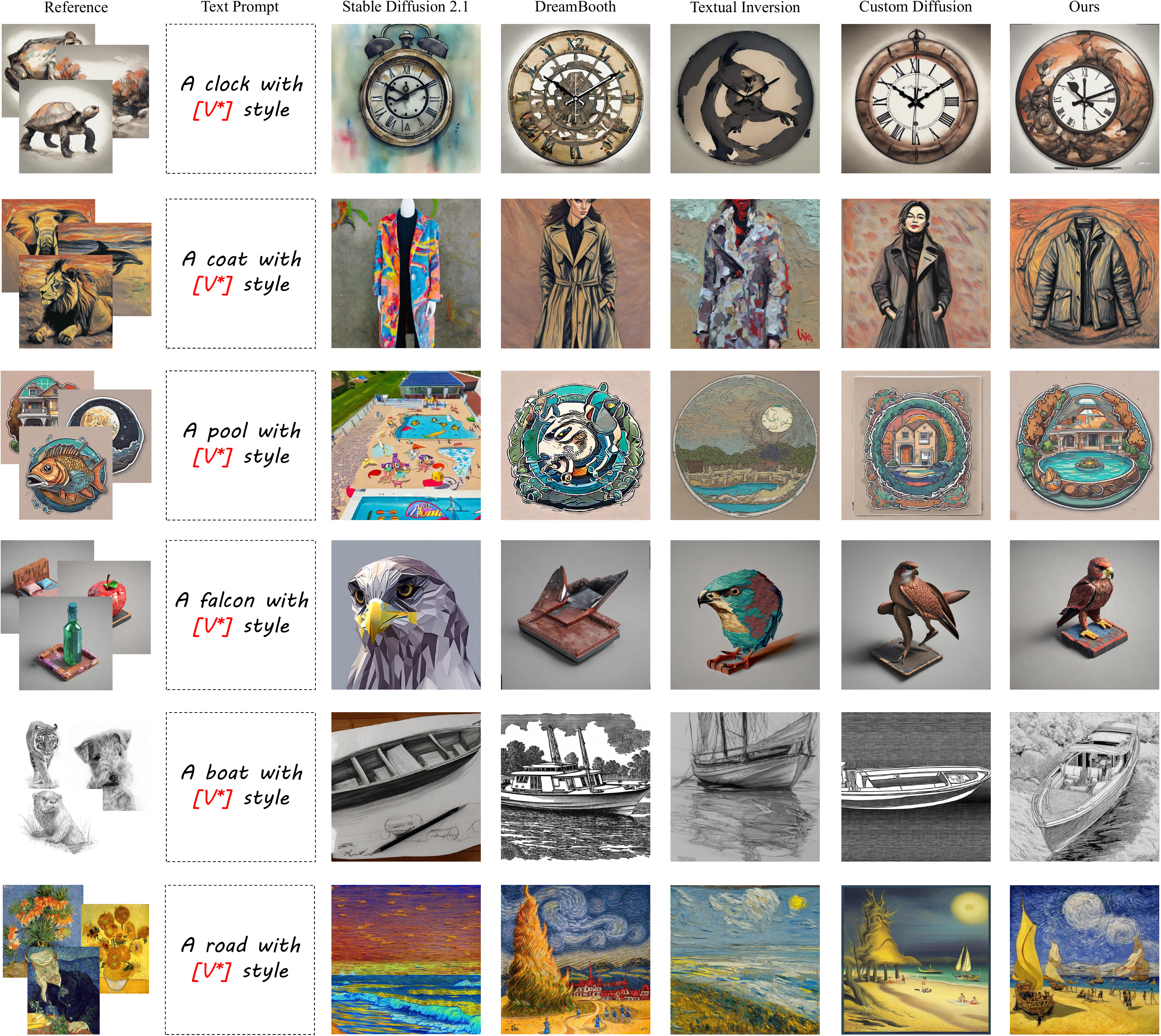}
  \caption{\textbf{Qualitative Performance Comparison.}}
  \label{Fig5}
\end{figure*}

\subsubsection{Style Keywords Reasoning}
To enable effective reverse style analysis, our approach constructs style-specific prompts tailored to extract stylistic attributes from input images. Inspired by prompt engineering techniques in recent works \cite{avrahami2023break, gao2024b, rombach2022high}, we leverage a Visual Language Model (e.g., GPT) to extract concise and descriptive style keywords from the style reference images. These prompts are designed to capture high-level stylistic features while avoiding ambiguity or irrelevant elements. During prompt construction, we utilize a structured input template to guide the Visual Language Model. Following prior works \cite{shi2024instantbooth, huang2023region}, the template explicitly focuses on stylistic and compositional elements present in the input image while excluding object-specific or contextual information. This ensures that the output is purely stylistic, enabling better disentanglement of style and content. The output is provided in JSON format for ease of integration into next steps. The Following is the template example used for this process.

Analyze the provided images, depicting visual style keywords. Extract and describe the stylistic attributes related to geometric patterns, material aesthetics, or artistic techniques, and summary all the feature in 1–3 concise descriptive keywords for each stylistic category. Avoid references to specific objects, colors, or contextual elements. Return the results in a dictionary format with a single key of ‘style keywords’ and one single result.
For examples, the answer might be:
    \{
      ``style keywords”: ``geometric reliefs"
    \}

This structured approach ensures the stability of large model outputs and guarantees the effectiveness and reliable performance of the generated content.

\subsubsection{Style Embeddings Initialization}
Following the above operations, the style keywords are encoded by CLIP as multi-dimensional embeddings \( E_{\text{gpt}} \), but the initial style identifier \( P_{\text{init}} \) (e.g., \( V^* \)) is encoded as one-dimensional embedding \( E_{\text{init}} \). To resolve dimensional mismatches, we expand the initial identifiers (e.g., \( V_1^*, V_2^* \)) to match the dimensions of \( E_{\text{gpt}} \), enabling accurate embedding alignment and replacement.

The initialized style embeddings are further refined by minimizing the Latent Diffusion Model (LDM) loss on sampled images, defined as:
\begin{equation}
\mathcal{L}_{\text{LDM}} = \mathbb{E}_{x,c,\epsilon,t} \left[ w_t \| G_{\text{style}}(\alpha_t x + \sigma_t \epsilon, c) - x \|^2_2 \right]
\label{eq:LDM}
\end{equation}
where \( G_{\text{style}} \) is a pre-trained diffusion model, \( c \) encodes style information, and \( \epsilon \) represents random Gaussian noise. \( \alpha_t, \sigma_t \) adjust noise influence at time \( t \), and \( w_t \) scales the loss.

The optimization objective for style embeddings \( V^* \) is:
\begin{equation}
V^* = \arg \min_{V^*} \mathbb{E}_{x, c, \epsilon, t} \left[ w_t \| G_{\text{style}}(\alpha_t x + \sigma_t \epsilon, c) - x \|^2_2 \right]
\end{equation}
This process focuses solely on optimizing style embeddings, keeping other model parameters freezed. By reusing the LDM training scheme, the embeddings effectively capture fine visual details associated with the target style, ensuring precise and detailed reconstruction.

\subsection{Joint Fine-tuning for Style Consistency and Content Preservation}  

In the process of fine-tuning text-to-image models, relying solely on the LDM loss often leads to issues such as content inconsistency and overfitting during style transfer. This happens because the LDM loss primarily focuses on reconstructing the input images, which may result in the model excessively prioritizing stylistic details while compromising the semantic alignment between the generated image and the input text prompt. To address this issue, we introduce a content preservation prior loss to ensure that the generated images maintain their core content attributes while integrating stylistic modifications, effectively balancing content fidelity and stylistic transformation. By introducing this loss, we aim to enhance the robustness and generalization capability of the model, even when training with limited samples.  

The content preservation prior loss is formulated as follows:  
\[
\mathcal{L}_{\text{content}} = \mathbb{E}_{B, c_{pr}, \epsilon', t'} \left[ w_{t'} \| G_{\text{content}}(\alpha_{t'} B + \sigma_{t'} \epsilon', c_{\text{pr}}) - B \|_2^2 \right]
\]  
Here, \( B \) represents the content reference images generated during data augmentation, which depict the same objects as the current reference images. \( G_{\text{content}} \) is the function that generates non-stylized images using the pre-trained model, ensuring that the content remains intact. \( \alpha_{t'} \) and \( \sigma_{t'} \) are time-step-related scaling factors, while \( \epsilon' \) is random Gaussian noise. The conditioning vector \( c_{\text{pr}} \) excludes style information, enabling the model to focus purely on content fidelity. The weighting factor \( w_{t'} \) determines the importance of the content preservation term. This loss term directly supervises the alignment between the non-stylized reference images and their generated counterparts, ensuring that the fundamental content attributes remain unchanged.  

In our joint fine-tuning approach, we utilize a combination of the LDM loss and the content preservation prior loss to optimize the model. The total loss function is expressed as:  
\[
\mathcal{L}_{\text{total}} = \lambda_1 \mathcal{L}_{LDM} + \lambda_2 \mathcal{L}_{\text{content}}
\]  
The first term, scaled by \(\lambda_1\) represents the LDM loss, As shown in Equation \ref{eq:LDM}. The second term, scaled by \( \lambda _2\), corresponds to the content preservation prior loss described above. By combining these two losses, our approach ensures that the model balances content consistency and stylistic adaptation during training.  

To further enhance the model’s ability to capture stylistic details without compromising content fidelity, we fine-tune both the multi-head attention layers in the U-Net and the Text Encoder. The U-Net primarily focuses on denoising and restoring fine-grained image details, such as textures and edges, which are crucial for realistic image generation. Meanwhile, the Text Encoder translates input text prompts into semantic vectors, capturing subtle distinctions such as specific shades or textures. Fine-tuning the Text Encoder enables the model to better interpret nuanced text descriptions and encode them into latent representations that guide image generation.  

By integrating the content preservation prior loss with the fine-tuning of U-Net and the Text Encoder, our method achieves a robust balance between style consistency and content fidelity. This joint optimization framework ensures that the generated images remain semantically aligned with the input text while accurately incorporating stylistic transformations, even in scenarios with limited training samples.

\begin{table}[t]
\caption{\textbf{Quantitative Results.} Bolded values indicate the best performance for each evaluation metric, while underlined values denote the second-best performance.}
\label{tab3}
\begin{center}
\renewcommand{\arraystretch}{1.1}
\resizebox{0.45\textwidth}{!}{  
\begin{tabular}{c|ccc}
    \toprule
    Methods & Pixel-Hist $\uparrow$ & CLIP R-Precision $\uparrow$ & CLIP-IQA $\uparrow$ \\  
    \midrule
    Stable Diffusion 2.1 \cite{rombach2022high} & 0.4736 & \textbf{0.8928} & 0.6136 \\
    DreamBooth \cite{ruiz2023dreambooth} & 0.6706 & 0.7168 & \underline{0.6199} \\
    Custom Diffusion \cite{kumari2023multi} & \underline{0.6765} & 0.7461 & 0.6137 \\
    Textual Inversion \cite{gal2022image} & 0.5345 & 0.5221 & 0.6067 \\
    Ours & \textbf{0.7392} & \underline{0.7714} & \textbf{0.6253} \\
    \bottomrule
\end{tabular}
}
\end{center}
\end{table}

\section{Experiments}
In this section, we present our experimental setup and findings, including the datasets used for training and evaluation, experimental settings with baseline models and parameter configurations, evaluation metrics for performance assessment, quantitative comparisons with existing methods, and an ablation study to analyze the contributions of different components of our approach.

\subsection{Datasets}

 \paragraph{Jittor Dataset} The Jittor Dataset\footnote{\url{https://www.educoder.net/competitions/index/Jittor-5}} consists of multiple categories of stylized images. For our study, we select 3 stylized images of each category for totally 28 categories for training, to evaluate the model's ability to capture stylistic nuances and generalize effectively in low-data scenarios.

 \paragraph{Style30K Dataset} Style30K Dataset \cite{li2024styletokenizer} is a large-scale dataset of 30,000 images with distinct stylistic features. Since our research focuses on style analysis, we refined a subset of 30 categories, Style30K-S, emphasizing style-related attributes for closer alignment with our objectives. While prior research on neural style transfer focuses mainly on painting styles \cite{gatys2015neural, deng2022stytr2}, we expand this scope to include a broader range of visual styles, enhancing its utility for style transfer applications.

\subsection{Experimental Setting}
\paragraph{Baselines}
We employ Stable Diffusion \cite{rombach2022high}, Textual Inversion \cite{gal2022image}, Custom Diffusion \cite{kumari2023multi}, and DreamBooth \cite{ruiz2023dreambooth} as baselines for personalizing text-to-image generation. Stable Diffusion \cite{rombach2022high} excels in generating high-quality, diverse, and controllable outputs. Textual Inversion \cite{gal2022image} enhances personalization by mapping text embeddings to visual features. Custom Diffusion \cite{kumari2023multi} improves image quality while maintaining stylistic coherence. DreamBooth \cite{ruiz2023dreambooth} enables effective fine-tuning with minimal data, producing consistent outputs aligned with specified styles.

\paragraph{Parameter Setting} In our experiments, all images were resized to 256×256 pixels, and the experiments were conducted on an NVIDIA 4090 GPU. The training process is divided into two stages. In the first stage, the model is trained for 500 steps with a learning rate of 1e-6 to refine the style embeddings. In the second stage, the model is fine-tuned for 2500 steps with a learning rate of 5e-5 to adapt both the U-Net and Text Encoder. The batch size is set to 1, and the entire optimization process takes approximately 5 minutes per style group.

\begin{figure*}
  \includegraphics[width=1\linewidth]{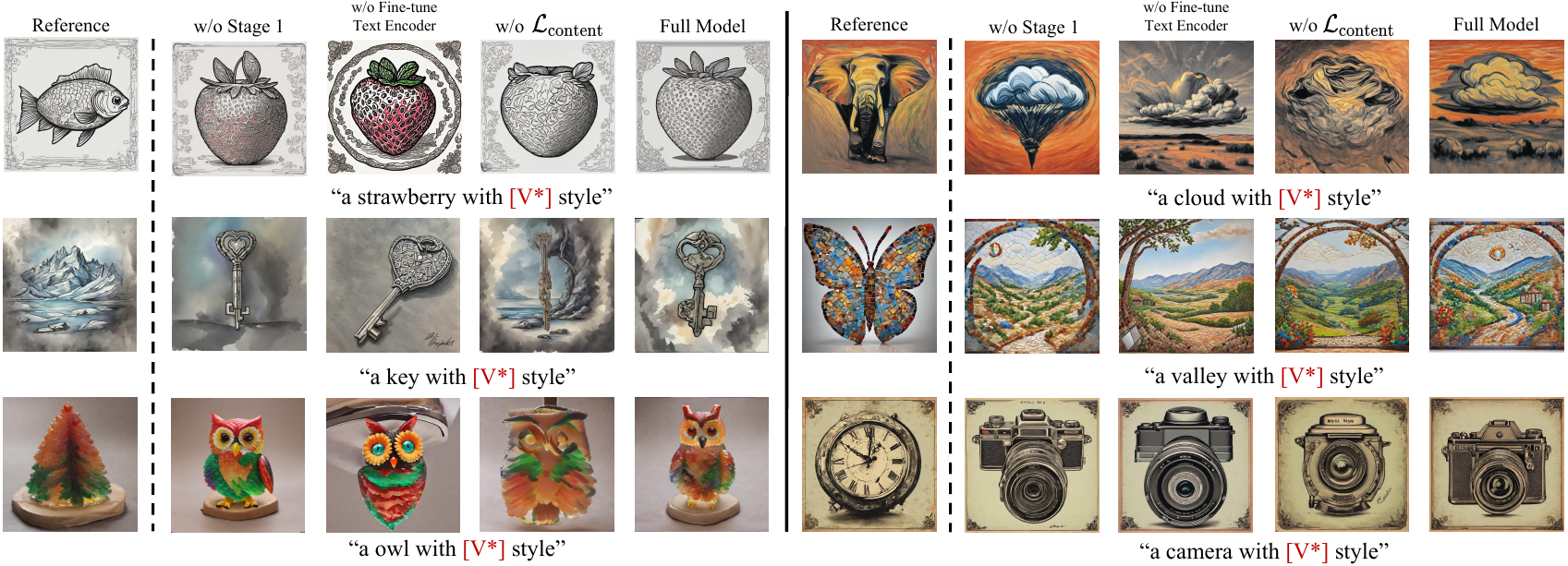}
  \caption{\textbf{Ablation Study.} 
  This figure presents ablation experiments, comparing results generated without our Stage I in Figure \ref{FigModel} (w/o Stage I), without Fine-tuning Text Encoder (w/o Fine-tune Text Encoder), without content preservation prior loss (w/o \(\mathcal{L}_{content}\)), and using our full method. The number of reference images is 3. The comparisons clearly demonstrate the superior performance and effectiveness of our complete approach in capturing stylistic attributes andf  preserving content fidelity.}
  \label{FigAblation}
\end{figure*}

\subsection{Evaluation Metrics}

We evaluate image quality and alignment from three perspectives: style, content, and quality. For style, we use Pixel-Hist to assess color distribution, where higher scores indicate more natural and visually appealing colors. For content, CLIP R-Precision \cite{park2021benchmark} measures semantic alignment between the generated images and input text, with higher scores reflecting better consistency. Finally, for quality, CLIP-IQA \cite{wang2023exploring} evaluates both visual appeal and semantic adherence, with higher scores denoting superior image quality.

\subsection{Comparison with Existing Methods}
\paragraph{Qualitative Comparison} 
We evaluate the performance of our method and baseline models using the prompt “an [obj] with [V*] style” across challenging style images from the Jittor and Style30K-S datasets. As shown in Figure \ref{Fig5}, our approach surpasses baseline models such as DreamBooth, Textual Inversion, Custom Diffusion, and Stable Diffusion with style keywords in maintaining stylistic consistency and ensuring accurate image content representation. Our method demonstrates superior text-image alignment and delivering higher stylization quality and detail preservation than other methods. These results highlight the robustness and effectiveness of our approach in personalized style transfer.

\paragraph{Quantitative Comparison} 
The quantitative results, presented in Table \ref{tab3}, validate the superiority of our method across key metrics such as Pixel-Hist, CLIP R-Precision, and CLIP-IQA. Our method outperforms baseline models in stylization similarity, content consistency, and image quality, achieving the highest Pixel-Hist and CLIP-IQA scores. Stable Diffusion with style prompt performs relatively well in CLIP R-Precision, though the influence of stylization has led to a decline in content metrics, our method achieves the best balance between style transfer quality and content fidelity, effectively demonstrating its capability in generating diverse, high-quality stylized outputs.

\subsection{Ablation Study}
\paragraph{Without Stage I} Stage I represents the style embeddings initialization based on VLM reasoning shown in Figure \ref{FigModel}. As illustrated in Figure \ref{FigAblation}, By removing this stage, although the model can still learn the style of the reference image, the stylization quality is significantly reduced, and the generated images lack detail, As shown in Figure \ref{FigAblation}. The absence of well-initialized embeddings results in uneven color gradients and poorly blended textures, diminishing the overall aesthetic quality. Furthermore, crucial texture details are often oversimplified or missing, while structural integrity is compromised, leading to visually unappealing and less accurate representations of the reference style. These observations emphasize the importance of Embeddings Initialization in achieving effective and high-quality style transfer.

\paragraph{Without Fine-tuning Text Encoder} As illustrated in Figure \ref{FigAblation}, we performed ablation experiments by removing Fine-tuning Text Encoder. The results reveal that without Fine-tuning Text Encoder, the model fails to effectively learn the style from the reference images, leading to compromised visual quality. Specifically, the model encounters difficulties in capturing essential stylistic features, such as color distribution, texture details, and structural patterns, resulting in inconsistent and incoherent style transfer. This deficiency manifests as blurry or distorted images with diminished detail and aesthetic fidelity. As demonstrated in Table \ref{tabCombined}, the absence of Text Encoder Training leads to a notable deterioration in metric of Pixel Hist, underscoring its critical role in personalized image style generation.

\paragraph{Without Content Preservation Prior Loss} As shown in Figure \ref{FigAblation}, while the model is capable of learning the style from the reference image, the alignment between the generated content and the text prompt noticeably decreases in the absence of content preservation prior loss. The model overly focuses on reproducing stylistic elements, resulting in reduced fidelity to the target subject and even overfitting. For example, deviations in content, such as unexpected elements or incoherence in object relationships, are observed. This highlights the necessity of content preservation prior loss in maintaining content fidelity while accurately transferring stylistic features.

\begin{table}[t]
\caption{\textbf{Ablation Study Results.} Bolded values indicate the best performance for each evaluation metric, while underlined values denote the second-best performance.}
\label{tabCombined}
\begin{center}
\footnotesize 
\resizebox{0.45\textwidth}{!}{  
\begin{tabular}{c|ccc}
    \toprule
    Methods & Pixel-Hist $\uparrow$ & CLIP R-Precision $\uparrow$ & CLIP-IQA $\uparrow$ \\  
    \midrule
    w/o Stage I & \underline{0.7281} & \underline{0.7741} & 0.6185 \\
    w/o Fine-tune Text Encoder & 0.6811 & \textbf{0.8381} & \textbf{0.6282} \\
    w/o \(\mathcal{L}_{content}\) & 0.7106 & 0.6890 & \underline{0.6269} \\
    Full Model & \textbf{0.7392} & 0.7714 & 0.6253 \\
    \bottomrule
\end{tabular}
}
\end{center}
\end{table}

\section{Conclusions}
We present a novel methodology for Personalizing Text-to-Image Style Generation, addressing key challenges in content consistency and precise representation of diverse artistic styles. Our approach significantly improving image stylization and text alignment. By introducing a style refinement strategy and integrating a content preservation prior loss function, we achieve a balance between stylistic coherence and content preservation, reducing overfitting and ensuring that generated outputs align closely with user preferences. Comprehensive experimental results confirm the effectiveness of our method, highlighting its superiority over existing models and its potential applications in personalized image generation.

\section*{Acknowledgment}

This work is supported in part by the Shandong Province Youth Entrepreneurship Technology Support Program for Higher Education Institutions, 2022KJN028, the Excellent Young Scientists Fund Program (Overseas) of Shandong Province (Grant No.2023HWYQ-034).

{\small
\nocite{yang2025curriculum, liu2022prompt, dong2024text, zheng2024unifying,  chen2024font, ma2024multimodal, ma2024plug, dong2022disentangled,chen2023class}
\bibliographystyle{IEEEtran} 
\bibliography{references}

}

\end{document}